\documentclass[10pt,twocolumn,letterpaper]{article}

\usepackage{iccv}
\usepackage{times}
\usepackage{epsfig}
\usepackage{graphicx}
\usepackage{amsmath}
\usepackage{amssymb}

\usepackage{booktabs}
\usepackage{makecell}
\usepackage{multirow}
\usepackage{eso-pic}

\newcommand{\methodname}{PanoOcc}

\usepackage[pagebackref=true,breaklinks=true,letterpaper=true,colorlinks,bookmarks=false]{hyperref}

\iccvfinalcopy % *** Uncomment this line for the final submission

\def\iccvPaperID{2662} % *** Enter the2 ICCV Paper ID here

\begin{document}

%%%%%%%%% TITLE
\title{PanoOcc: Unified Occupancy Representation for Camera-based \\ 3D Panoptic Segmentation}

\author{
  Yuqi Wang$^{1,2}$\quad 
  Yuntao Chen$^{3}$\quad 
  Xingyu Liao$^\ast$ \quad 
  Lue Fan$^{1}$ \quad
  Zhaoxiang Zhang$^{1,2,3}$ \quad
  \\ 
$^1$ CRIPAC, Institute of Automation, Chinese Academy of Sciences (CASIA)\\
$^2$ School of Artificial Intelligence, University of Chinese Academy of Sciences (UCAS) \\
$^3$ Centre for Artificial Intelligence and Robotics, HKISI, CAS \\
  \tt\small{\{wangyuqi2020,fanlue2019,zhaoxiang.zhang\}@ia.ac.cn}  \quad \tt\small{chenyuntao08@gmail.com} \\
  \tt\small{randall@mail.ustc.edu.cn} \\
}

\maketitle{
\let\thefootnote\relax\footnote{$\ast$ Independent researcher
}

%%%%%%%%% ABSTRACT
\begin{abstract}
   Comprehensive modeling of the surrounding 3D world is key to the success of autonomous driving.
However, existing perception tasks like object detection, road structure segmentation, depth \& elevation estimation, and open-set object localization each only focus on a small facet of the holistic 3D scene understanding task.
This divide-and-conquer strategy simplifies the algorithm development procedure at the cost of losing an end-to-end unified solution to the problem.
In this work, we address this limitation by studying \textbf{camera-based 3D panoptic segmentation}, aiming to achieve a unified occupancy representation for camera-only 3D scene understanding.
To achieve this, we introduce a novel method called \textbf{\methodname{}}, which utilizes voxel queries to aggregate spatiotemporal information from multi-frame and multi-view images in a coarse-to-fine scheme, integrating feature learning and scene representation into a unified occupancy representation.
We have conducted extensive ablation studies to verify the effectiveness and efficiency of the proposed method.
Our approach achieves new state-of-the-art results for camera-based semantic segmentation and panoptic segmentation on the nuScenes dataset.
Furthermore, our method can be easily extended to dense occupancy prediction and has shown promising performance on the Occ3D benchmark.
The code will be released at \url{https://github.com/Robertwyq/\methodname}.

\end{abstract}

%%%%%%%%% BODY TEXT
\section{Introduction}

% 3D scene understanding is important
Holistic 3D scene understanding is vital in autonomous driving.
The capability to perceive the environment, identify and categorize objects, and contextualize their positions in the 3D space of the scene is fundamental for developing a safe and reliable autonomous driving system.

\begin{figure}[t]
  \centering
    \includegraphics[width=1.0\linewidth]{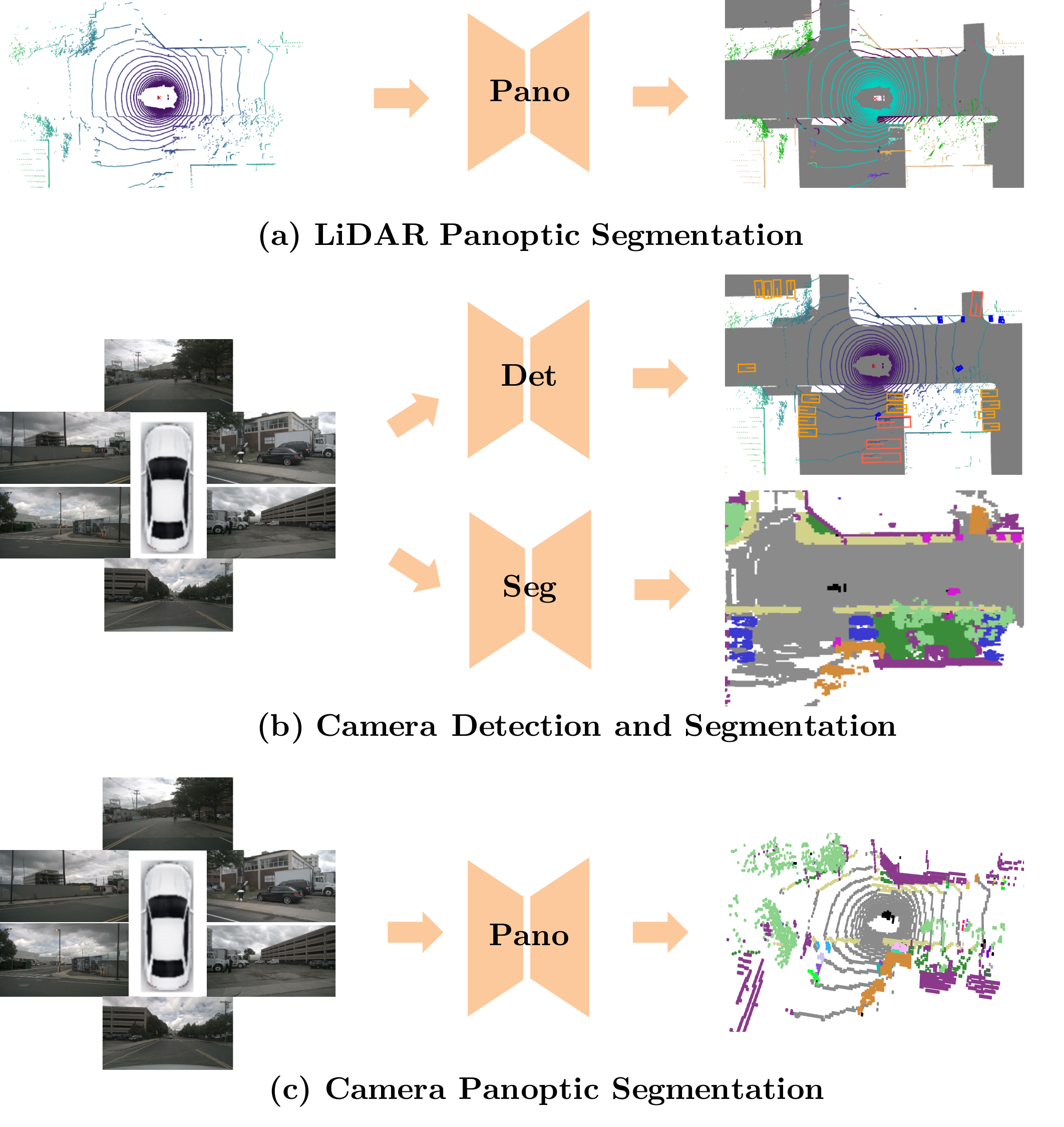}
    \caption{\textbf{Comparison of different tasks for 3D scene understanding.} (a) LiDAR panoptic segmentation: Given sparse LiDAR points as input, the model outputs panoptic prediction on sparse LiDAR points. (b) Camera Detection and Segmentation: Given multi-view images, separate models are used to detect objects and perform BEV semantic segmentation. (c) Camera panoptic segmentation: Given multi-view images, a single model is trained to output dense panoptic occupancy predictions.}
  \label{fig:camera_panoptic}
  \vspace{-12pt}
\end{figure}

% unified task (multi-task, output representation)
Recent advancements in camera-based Bird's Eye View (BEV) methods have shown great potential in enhancing 3D scene understanding. By integrating multi-view observations into a unified BEV space, these methods have achieved remarkable success in tasks such as 3D object detection~\cite{wang2022detr3d,li2022bevformer,liu2022petrv2,li2022bevdepth}, BEV semantic segmentation~\cite{philion2020lift,hu2021fiery,zhou2022cross}, and vector map construction~\cite{liu2022vectormapnet,liao2022maptr}.
However, existing perception tasks have certain limitations as they primarily focus on specific aspects of the scene. Object detection is primarily concerned with identifying foreground objects, BEV semantic segmentation only predicts the semantic map on the BEV plane, and vector map construction emphasizes the static road structure of the scene.
To address these limitations, there is a need for a more comprehensive and integrated paradigm for 3D scene understanding.
In this paper, we propose \emph{camera-based panoptic segmentation}, which aims to encompass all the elements within the scene in a unified representation for the 3D output space. As shown in Figure~\ref{fig:camera_panoptic}, unlike LiDAR-based panoptic segmentation (a) that relies on LiDAR point clouds, our camera-based panoptic segmentation leverages multi-view images as input and outputs a dense panoptic occupancy prediction throughout the entire scene. In contrast to recent camera-based detection and segmentation methods (b), it seamlessly integrates object-level and voxel-level perception results into a unified panoptic occupancy representation.

% unified feature learning
However, directly utilizing Bird's Eye View (BEV) features for camera-based panoptic segmentation leads to poor performance due to the omission of finer geometry details, such as shape and height information, which are crucial for decoding fine-grained 3D structures. This limitation motivates us to explore a more effective 3D feature representation.
Occupancy representation has gained popularity as it effectively describes various elements in the scene, including open-set objects (e.g., debris), irregular-shaped objects (e.g., articulated trailers, vehicles with protruding structures), and special road structures (e.g., construction zones). 
Therefore, a burst of recent methods~\cite{cao2022monoscene,huang2023tri, miao2023occdepth, cao2022monoscene,wang2023openoccupancy,li2023voxformer} have focused on dense semantic occupancy prediction.
However, simply lifting 2D to 3D occupancy representation has been considered inefficient in terms of memory cost. This limitation has driven methods like TPVFormer~\cite{huang2023tri} to split the 3D representation into three 2D planes. Although these methods attempt to mitigate the memory issue, they still struggle to capture the complete 3D information and may experience reduced performance.
Moreover, these existing works primarily focus on the semantic understanding of the scene and do not address instance-level discrimination.

% our work and contribution
In this work, we propose a novel method called \emph{\methodname{}}, which seamlessly integrates object detection and semantic segmentation in a joint-learning framework, facilitating a more comprehensive comprehension of the 3D environment. Both detection and segmentation performance can benefit from this joint-learning framework.
Our approach employs voxel queries to learn a unified occupancy representation.
This occupancy is learned in a coarse-to-fine scheme, solving the problem of memory cost and significantly enhancing efficiency.
We then take a step further to explore the sparse nature of 3D space and propose an occupancy sparsify module. This module progressively prunes occupancy to a spatially sparse representation during the coarse-to-fine upsampling, greatly boosting memory efficiency.
Our contributions are summarized as follows:
\begin{itemize}
    \item We introduce \emph{camera-based 3D panoptic segmentation} as a new paradigm for holistic 3D scene understanding, which utilizes multi-view images to create a unified occupancy representation for the 3D scene. This allows us to jointly model object detection and semantic segmentation, leading to a more cohesive and holistic understanding of the scene.
    \item Our proposed framework, \methodname{}, adopts a \emph{coarse-to-fine scheme} to learn the unified occupancy representation from multi-frame and multi-view images. We demonstrate that using 3D voxel queries with a coarse-to-fine learning scheme is effective and efficient.
    This scheme could be further made spatially sparse to boost memory efficiency by an occupancy sparsify module.
    \item Experiments on the nuScenes dataset show that our approach achieves state-of-the-art performance on camera-based semantic segmentation and panoptic segmentation.
    Furthermore, our approach can extend to dense occupancy prediction and has shown promising performance on the Occ3D benchmark.
\end{itemize}

%-------------------------------------------------------------------------
\section{Related Work}
\label{sec:related}

\noindent{\textbf{Camera-based 3D Perception.}}
Camera-based 3D perception has received extensive attention in the autonomous driving community due to its cost-effectiveness and rich visual attributes.
Previous methods perform 3D object detection and map segmentation tasks independently. Recent BEV-based methods unify these tasks on the problem of feature view transformation from image space to BEV space.
One line of works follows the lifting paradigm proposed in LSS~\cite{philion2020lift}; they explicitly predict a depth map and lift multi-view image features onto the BEV plane~\cite{huang2021bevdet,li2022bevdepth,li2022bevstereo,park2022time}.
Another line of works inherits the spirit of querying from 3D to 2D in DETR3D~\cite{wang2022detr3d}; they employ learnable queries to extract information from image features by cross-attention mechanism~\cite{li2022bevformer,lu2022learning,jiang2022polarformer,wang2023frustumformer}.
While these methods efficiently compress information onto the BEV plane, they may sacrifice some of the integral scene structure inherent in 3D space. To address this limitation, voxel representation is better suited for obtaining a holistic understanding of 3D space, making it ideal for tasks such as 3D semantic segmentation and panoptic segmentation.

\noindent{\textbf{3D Occupancy Prediction.}}
Occupancy prediction can be traced back to Occupancy Grid Mapping (OGM)~\cite{thrun2002probabilistic}, a classic task in mobile robot navigation that aims to generate probabilistic maps from sequential noisy range measurements. Recently, there has been considerable attention given to camera-based 3D occupancy prediction, which aims to reconstruct the 3D scene structure from images. Existing tasks in this area can be categorized into two lines based on the type of supervision: sparse prediction and dense prediction.
Sparse prediction methods obtain supervision from LiDAR points and are evaluated on LiDAR benchmarks. TPVFormer~\cite{huang2023tri} proposes a tri-perspective view method for predicting 3D occupancy.
Dense prediction methods are closely related to Semantic Scene Completion (SSC)~\cite{armeni2017joint,song2017semantic,dai2017scannet,liao2022kitti}. 
MonoScene~\cite{cao2022monoscene} first uses U-Net to infer dense 3D occupancy with semantic labels from a single monocular RGB image. VoxFormer~\cite{li2023voxformer} utilizes depth estimation to select voxel queries in a two-stage framework. Subsequently, a series of studies have focused on the task of dense occupancy prediction and have introduced new benchmarks. OpenOccupancy~\cite{wang2023openoccupancy} provides a carefully designed occupancy benchmark, while Occ3D~\cite{tian2023occ3d} proposes an occupancy prediction benchmark using the Waymo and nuScenes datasets. Openocc~\cite{tong2023scene} further provides occupancy flow annotation on the nuScenes dataset.

\noindent{\textbf{LiDAR Panoptic Segmentation.}}
LiDAR panoptic segmentation~\cite{milioto2020lidar} offers a comprehensive understanding of the environment by unifying semantic segmentation and object detection. However, traditional object detection methods often lose height information, making it challenging to learn fine-grained feature representations for accurate 3D segmentation. Recent LiDAR panoptic methods~\cite{zhou2021panoptic,razani2021gp,hong2021lidar} have been developed based on well-designed semantic segmentation networks~\cite{zhang2020polarnet,cheng20212} to address this limitation.
Instead of predicting sparse semantic segmentation on LiDAR points, camera-based panoptic segmentation aims to output dense voxel segmentation of the scene.

\noindent{\textbf{3D Scene Reconstruction and Representation.}}
3D scene reconstruction and representation aim to infer the holistic geometry structure and semantics of a scene. This challenging problem has received continuous attention in both the traditional computer vision era and the recent deep learning era~\cite{han2019image}. Solutions can be categorized into \emph{explicit reconstruction} and \emph{implicit representation} approaches.
Explicit reconstruction leverage the geometry cues from different viewpoints in multi-views~\cite{schonberger2016structure,schonberger2016pixelwise}. 
While explicit reconstruction methods excel at reconstructing static scenes, they are struggled to capture dynamic scenes or scenes with complex interactions between objects. Furthermore, they are computationally expensive, requiring large amounts of time to generate detailed and accurate 3D models.
In contrast, implicit representation methods are more computation efficient and have the potential to model scenes at arbitrary resolutions. They learn a continuous function~\cite{mescheder2019occupancy,park2019deepsdf,mildenhall2021nerf,sitzmann2019deepvoxels} that can represent complex 3D scenes with high fidelity, including hidden or occluded regions that are difficult to capture using explicit reconstruction methods.
% Our camera-based panoptic segmentation is more like a hybrid representation~\cite{chen2022tensorf} for scene understanding, introducing explicit geometry projections into the voxel representation learning.

%-------------------------------------------------------------------------
\section{Methodology}
\subsection{Problem Setup}
\begin{figure*}
  \centering
    \includegraphics[width=1.0\linewidth]{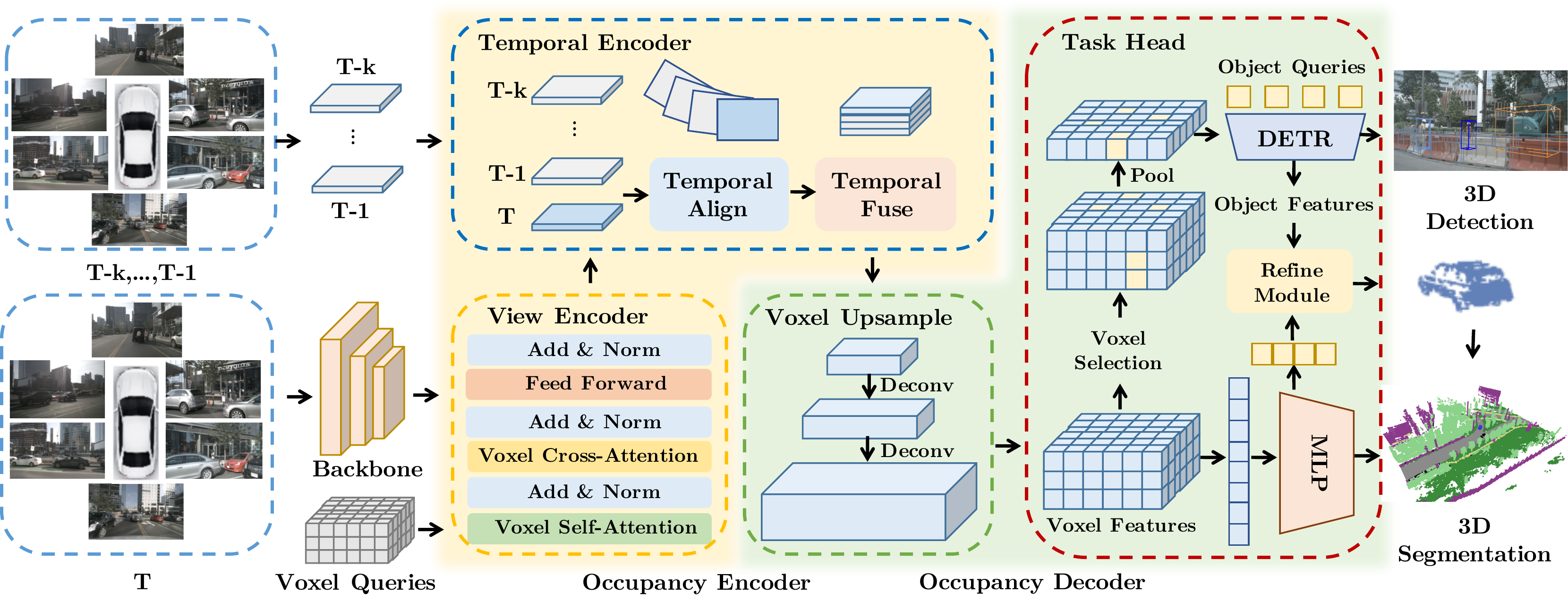}
  \caption{\textbf{The overall framework of \methodname{}.} We employ an image backbone network to extract multi-scale features for multi-view images at multi-frames. Then we apply voxel queries to learn the voxel features via \emph{View Encoder}. The \emph{Temporal Encoder} aligns the previous voxel features into the current frame and fuses the features together. \emph{Voxel Upsample} restores the high-resolution voxel representation for fine-grained semantic classification. \emph{Task Head} predicts object detection and semantic segmentation by two separate heads. \emph{Refine Module} further refines the thing class prediction with the help of 3D object detection and assigns the instance ID to the thing-occupied grids. Finally, we can obtain 3D panoptic segmentation for the current frame.}
  \vspace{-10pt}
  \label{fig:pipeline}
\end{figure*}

\noindent{\textbf{Camera-based 3D panoptic segmentation.}}
Camera-based 3D panoptic segmentation aims to predict a dense panoptic voxel volume surrounding the ego-vehicle using multi-view images as input.
Specifically, we take current multi-view images denoted as $\mathbf{I}_t=\{\mathbf{I}_t^{1}, \mathbf{I}_t^{2},...,\mathbf{I}_t^{n}\}$ and previous frames $\mathbf{I}_{t-1},...,\mathbf{I}_{t-k}$ as input. $n$ denotes the camera view index, while $k$ denotes the number of history frames.
The model outputs the current frame semantic voxel volume $\mathbf{Y}_t \in \{w_0, w_1,...,w_C\}^{ H \times W \times Z}$ and its corresponding instance ID $\mathbf{N}_t \in \{v_0, v_1, v_2,...,v_P\}^{ H \times W \times Z}$.
Here $C$ denotes the total number of semantic classes in the scene, while $w_0$ represents the empty voxel grid. $P$ are the total number of instances in the current frame $t$; for each grid belonging to the foreground classes (\emph{thing}), it would assign a specific instance ID $v_j$. $v_0$ is assigned to all voxel grids belonging to the \emph{stuff} and empty.
$H, W, Z$ denotes the length, width, and height of the voxel volume.

\noindent{\textbf{Camera-based 3D semantic occupancy prediction.}
Camera-based 3D semantic occupancy prediction can be considered a sub-problem of camera-based 3D panoptic segmentation. The former focus only on predicting the semantic voxel volume $\mathbf{Y}_t \in \{w_0, w_1,...,w_C\}^{ H \times W \times Z}$.
However, the emphasis is particularly placed on accurately distinguishing the empty class ($w_0$) from the other classes to determine whether a voxel grid is empty or occupied. 

%-------------------------------------------------------------------------------
\subsection{Overall Architecture}
In this section, we introduce the overall architecture of \methodname{}. As shown in Figure~\ref{fig:pipeline}, our method takes multi-frame multi-view images as input and first extracts multi-scale features using an image backbone. 
These features are then processed by the \emph{Occupancy Encoder}, which consists of the \emph{View Encoder} and \emph{Temporal Encoder}, to generate a coarse unified occupancy representation. The \emph{View Encoder} utilizes voxel queries to learn voxel features, preserving the actual 3D structure of the scene by explicitly encoding height information. The \emph{Temporal Encoder} aligns and fuses previous voxel features with the current frame, capturing temporal information and enhancing the representation.
Next, the \emph{Occupancy Decoder} employs a coarse-to-fine scheme to recover fine-grained occupancy representation. The \emph{Coarse-to-fine Upsampling} module restores the high-resolution voxel representation, enabling precise semantic classification. The \emph{Task Head} predicts object detection and semantic segmentation using separate heads.
To refine the prediction of \emph{thing} classes, the \emph{Refine Module} leverages 3D object detection results and assigns instance IDs to the thing-occupied grids.
By combining these modules, our method produces 3D panoptic segmentation for the current scene. In the following sections, we provide detailed descriptions of each module.

\subsection{Voxel Queries}
We define a group of 3D-grid-shape learnable parameters $\mathbf{Q} \in \mathbb{R}^{H\times W\times Z \times D}$ as voxel queries.
$H$ and $W$ are the spatial shape of the BEV plane, while $Z$ represents the height dimension.
A single voxel query $\mathbf{q} \in \mathbb{R}^{D}$ located at $(i, j, k)$ position of $\mathbf{Q}$ is responsible for the corresponding 3D voxel grid cell region. Each grid cell in the voxel corresponds to a real-world size of $(s_h,s_w, s_z)$ meters.
To incorporate positional information in the voxel queries, we add learnable 3D positional embeddings to $\mathbf{Q}$.

\subsection{Occupancy Encoder}
Given voxel queries $\mathbf{Q}$ and extracted image feats $\mathbf{F}$ as input, the occupancy encoder outputs the fused voxel features $\mathbf{Q}_f \in \mathbb{R}^{H\times W\times Z \times D}$. $H$,$W$ and $Z$ represent the shape of the output voxel features, and $D$ is the embedding dimension.

\noindent{\textbf{View Encoder.}}
View Encoder transforms the perspective view features into 3D voxel features. 
However, applying vanilla cross-attention for this view transformation can be computationally expensive when dealing with voxel representation.
To address this issue, we draw inspiration from the querying paradigm in recent BEV-based methods~\cite{li2022bevformer,jiang2022polarformer,wang2023frustumformer} and adopt efficient deformable attention~\cite{zhu2020deformable} for voxel cross-attention and voxel self-attention. 
The core difference lies in the choice of \emph{reference points} to generalize the BEV queries to voxel queries.

The voxel cross-attention is designed to facilitate the interaction between multi-scale image features and voxel queries. Specifically, for a voxel query $\mathbf{q}$ located at $(i, j, k)$, the process of voxel cross-attention (VCA) can be formulated as follows:
\begin{equation}
    \text{VCA}(\mathbf{q},\mathbf{F}) = \frac{1}{|v|}\sum_{n\in v}\sum_{m=1}^{M_1} \text{DA} (\mathbf{q}, \mathbf{\pi_n}(\mathbf{Ref}_{i,j,k}^{m}),\mathbf{F}_n)
\end{equation}
where $n$ indexes the camera view, $m$ indexes the reference points, and $M_1$ is the total number of sampling points for each voxel query. $v$ is the set of image views for which the projected 2D point of the voxel query can fall on.
$\mathbf{F}_{n}$ is the image features of the $n$-th camera view. 
$\mathbf{\pi_n}(\mathbf{Ref}_{i,j,k}^{m})$ denotes the $m$-th projected reference point in $n$-th camera view, projected by projection matrix $\mathbf{\pi_n}$ from the voxel grid located at $(i, j, k)$. $\text{DA}$ represents deformable attention.
The real position of a reference point located at voxel grid $(i, j, k)$ in the ego-vehicle frame is $(x_i^m,y_j^m,z_k^m)$. 
The projection between $m$-th projected reference point $\mathbf{Ref}_{i,j,k}^{m}$ and its corresponding 2D reference point $(u_{ijk}^{n,m},v_{ijk}^{n,m})$ on the $n$-th view can be formulate as:
\begin{equation}
    \mathbf{Ref}_{i,j,k}^{m} = (x_i^m,y_j^m,z_k^m)
\end{equation}
\begin{equation}
    d_{ijk}^{n,m} \cdot [u_{ijk}^{n,m},v_{ijk}^{n,m},1] = \mathbf{P_n} \cdot [x_i^m,y_j^m,z_k^m,1]^T
\end{equation}
where $\mathbf{P}_n \in \mathbb{R}^{3\times 4}$ is the projection matrix of the $n$-th camera. $(u_{ijk}^{n,m},v_{ijk}^{n,m})$ denotes the $m$-th 2D reference point on $n$-th image view. $d_{ijk}^{n,m}$ is the depth in the camera frame.

Voxel self-attention (VSA) facilitates the interaction between voxel queries. For a voxel query $\mathbf{q}$ located at $(i, j, k)$, it only interacts with the voxel queries at the reference points nearby. The process of voxel self-attention can be formulated as follows:
\begin{equation}
    \text{VSA}(\mathbf{q},\mathbf{Q})= \sum_{m=1}^{M_2}\text{DA}(\mathbf{q},\mathbf{Ref}_{i,j,k}^{m},\mathbf{Q})
\end{equation}
where $m$ indexes the reference points, and $M_2$ is the total number of reference points for each voxel query. $\text{DA}$ represents deformable attention.
Contrary to the reference points on the image plane in voxel cross-attention, $\mathbf{Ref}_{i,j,k}^{m}$ in voxel self-attention is defined on the BEV plane.
\begin{equation}
    \mathbf{Ref}_{i,j,k}^{m} = (x_i^m,y_j^m,z_k)
\end{equation}
where $(x_i^m,y_j^m,z_k)$ denotes the $m$-th reference point for query $\mathbf{q}$. These sampling points share the same height $z_k$, but with different learnable offsets for $(x_i^m,y_j^m)$. This encourages the voxel queries to interact in the BEV plane, which contains more semantic information.

\noindent{\textbf{Temporal Encoder.}}
Temporal information is crucial in camera-based perception systems to understand the surrounding environment.
Recent breakthroughs in camera-based perception systems, such as BEV-based detectors~\cite{park2022time,li2022bevformer,liu2022petrv2}, have shown that incorporating temporal information can significantly improve the performance. 
We designed a temporal encoder adapted to the 3D voxel queries to further enhance the voxel representation. 

Temporal encoder incorporates the history voxel queries information ($\mathbf{Q}_{t-k},...,\mathbf{Q}_{t-1}$) to the current voxel queries $\mathbf{Q}_{t}$.
As shown in Figure~\ref{fig:pipeline}, the temporal encoder consists of two specific operations: \emph{temporal align} and \emph{temporal fuse}.
Different from previous temporal alignment methods~\cite{li2022bevformer,park2022time}, which align history features on the BEV plane, our approach employs voxel alignment in 3D space.
This allows us to correct for the inaccuracies caused by the assumptions made in previous BEV-based methods. They assumed that road height remains unchanged throughout the scene, which is not always valid in real-world driving scenarios, particularly when encountering uphill and downhill terrain.
Voxel alignment is crucial for fine-grained voxel representations to perceive the environment accurately. 
Specifically, the process of voxel alignment is formulated as follows:
\begin{equation}
    \mathbf{G}_{t-k} = \mathbf{T}_{t\rightarrow t-k}\cdot \mathbf{G}_t
\end{equation}
\begin{equation}
    \mathbf{Q}_{t-k \rightarrow t} = \text{GridSample}(\mathbf{Q}_{t-k}, \mathbf{G}_{t-k})
\end{equation}
where $\mathbf{G}_t \in \mathbb{R}^{H\times W\times Z}$ is the voxel grid at current frame $t$, $\mathbf{G}_{t-k} \in \mathbb{R}^{H\times W\times Z}$ represents the current frame grid at frame $t-k$. $\mathbf{T}_{t\rightarrow t-k}$ is the transformation matrix for transforming the points at frame $t$ to previous frame $t-k$. 
Then the queries at frame $t-k$ are aligned to current frame $t$ by interpolation sampling, denoted as $\mathbf{Q}_{t-k \rightarrow t}$.
After the alignment, the previous aligned voxel queries $[\mathbf{Q}_{t-k \rightarrow t},..., \mathbf{Q}_{t-1 \rightarrow t}]$ are concated with the current voxel queries $\mathbf{Q}_t$. We employ a block of residual 3D convolution to fuse the queries and output fused voxel queries $\mathbf{Q}_f$.

\subsection{Occupancy Decoder}
Given the coarse voxel feature $\mathbf{Q}_f$, it should be converted to a fine-grained feature to meet the need for panoptic occupancy prediction.

\noindent{\textbf{Coarse-to-fine Upsampling.}}
This module upsamples the fused voxel query $\mathbf{Q}_{f} \in \mathbb{R}^{H\times W\times Z\times D}$ to the high resolution occupancy features $\mathbf{O}  \in \mathbb{R}^{H'\times W'\times Z'\times D'}$ by 3D deconvolutions.
Such a coarse-to-fine manner not only avoids directly applying expensive 3D convolutions to high-resolution occupancy features, but also leads to no performance loss. We have a quantitative discussion in the experiment section.

\noindent{\textbf{Occupancy Sparsify.}}
Although the coarse-to-fine manner guarantees the high efficiency of our method, there is a considerable computational waste on the spatially dense feature $\mathbf{Q}_f$ and $\mathbf{O}$.
This is because our physical world is essentially sparse in spatial dimensions, which means a large portion of space is not occupied.
Dense operations (i.e., dense convolution) violate such essential sparsity.
Inspired by the success of sparse architecture in LiDAR-based perception~\cite{yan2018second, liu2022spatial, fan2023super}, we optionally turn to the Sparse Convolution~\cite{sparseconv} for occupancy sparsify.
In particular, we first learn an occupancy mask for $\mathbf{Q}_f$ to indicate if positions on $\mathbf{Q}_f$ are occupied.
Then we prune $\mathbf{Q}_f$ to a sparse feature $\mathbf{Q}_{sparse} \in \mathbb{R}^{N\times D}$ by discarding those empty positions according to the learned occupancy mask, where $N \ll HWZ$ and $N$ is determined by a predefined keeping ratio $R_{keep}$.
After the pruning, all the following dense convolutions are replaced by corresponding sparse convolutions.
Since sparse deconvolution will dilate the sparse features to empty positions and reduce the sparsity, we conduct similar pruning operations after each upsampling to maintain the spatial sparsity. 
Finally, we obtain a high-resolution and sparse occupancy feature $\mathbf{O}_{sparse}  \in \mathbb{R}^{N' \times D'}$, where $N' \ll H'W'Z'$.
Figure~\ref{fig:pruning} illustrates the occupancy sparsify process.
% \vspace{-10pt}
\begin{figure}[h]
  \centering
    \includegraphics[width=1.0\linewidth]{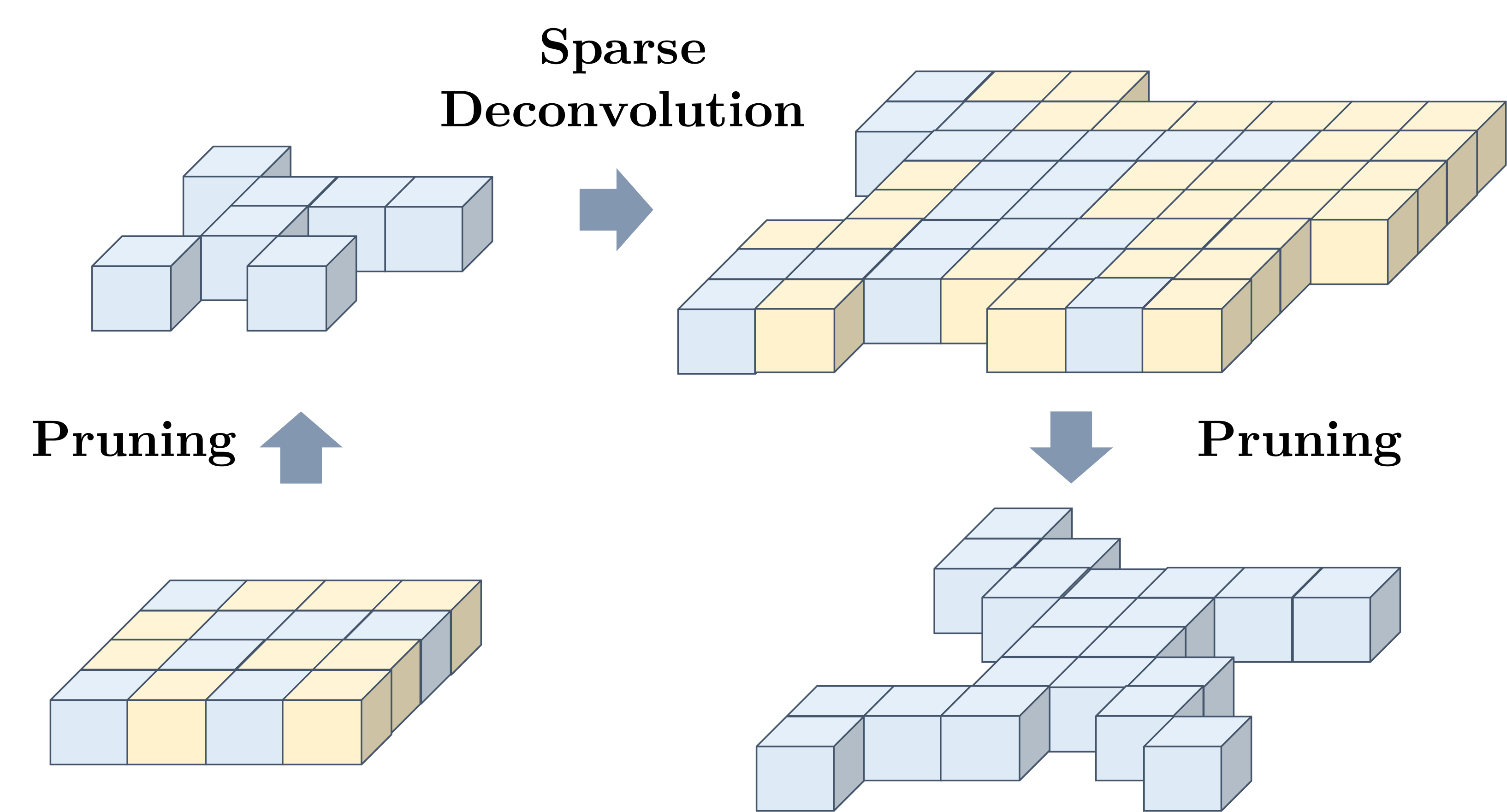}
    \caption{\textbf{Illustration of occupancy sparsify.}
    It serves as an optional technique to boost efficiency. We use BEV representation for simple illustration, while it is actually a 3D process.
    The light yellow region will be pruned according to occupancy masks.}
  \label{fig:pruning}
  \vspace{-10pt}
\end{figure}

\subsection{Multi-task Training}
Based on the unified occupancy representation, our model has a strong capacity to handle different tasks.
Specifically, our model is trained end-to-end for joint detection and segmentation, achieving the purpose of panoptic perception.

\noindent{\textbf{Voxel Selection.}}
Considering the detection task cares more about foreground classes (\emph{thing}), whereas the segmentation task must take into account all classes (\emph{stuff} and \emph{thing}). Because of the conflicting learning objectives, distinct features are required. Hence, we learn a binary voxel mask $\mathbf{M}$ to pick out the foreground voxel features for the detection part.
% The learning of voxel mask $\mathbf{M}$ is supervised by $\mathcal{L}_{thing}$.
The 2D BEV features are obtained by average pooling on the height dimension: $\mathbf{B} = Pool_{avg}(\mathbf{M}\cdot \mathbf{O})$.

\noindent{\textbf{Detection Head.}} 
Following ~\cite{li2022bevformer}, we adopt a query-based deformable-DETR head as the detection head.
The detection head is applied on the 2D BEV features $\mathbf{B}$.

\noindent{\textbf{Segmentation Head.}}
We employ a lightweight multilayer perceptron (MLP) head for semantic segmentation, based on occupancy feature $\mathbf{O}$ or the sparse counterpart $\mathbf{O}_{sparse}$.
This allows us to query the voxel grid status at arbitrary positions.

\noindent{\textbf{Losses.}}
The total loss $\mathcal{L}$ has two parts:
\begin{equation}
    \mathcal{L} = \mathcal{L}_{Det} + \mathcal{L}_{Seg}
\end{equation}
The voxel segmentation head is supervised by $\mathcal{L}_{Seg}$, a dense loss consisting of focal loss~\cite{lin2017focal} (all voxels) and Lovasz loss~\cite{berman2018lovasz} (voxels containing LiDAR points) for voxel prediction.
For Voxel Selection, we predict a binary voxel mask to select the foreground classes (\emph{thing}) voxel features for the object detection head, and the voxel mask is supervised by focal loss~\cite{lin2017focal}. The total loss $\mathcal{L}_{Seg}$ is formulated as:
\begin{equation}
    \mathcal{L}_{Seg} = \lambda_1 \mathcal{L}_{focal} + \lambda_2 \mathcal{L}_{lovasz} + \lambda_3 \mathcal{L}_{thing}
\end{equation}
The detection head is supervised by $\mathcal{L}_{Det}$, a sparse loss consisting of focal loss~\cite{lin2017focal} for classification and L1 loss for bounding box regression:
\begin{equation}
    \mathcal{L}_{Det} = \lambda_4 \mathcal{L}_{cls} + \lambda_5 \mathcal{L}_{reg}
\end{equation}

\noindent{\textbf{Refine Module.}} 
In this module, we refine the predicted foreground (\emph{thing}) voxels using the detection results.
We start by sorting all box predictions based on their confidence scores.
Then, we select a set of high-confidence bounding boxes denoted as $G = \{b_i|s_i>\tau\}$, where $b_i$ represents a 3D bounding box, $s_i$ is the confidence score, and $\tau$ is a threshold (default: $\tau=0.8$).
For the voxels within each bounding box $b_i$, we assign the class prediction $c_i$ to all of them. This improves segmentation consistency and slightly enhances the mean Intersection over Union (mIoU) by 0.1-0.2 points.
To perform panoptic voxel segmentation, we assign instance IDs sequentially based on confidence scores. If the current instance overlaps with previous instances beyond a certain threshold, we ignore it to avoid duplication.
Finally, we assign instance ID 0 to all voxels corresponding to the \emph{stuff} class.

%-------------------------------------------------------------------------
\section{Experiment}

\subsection{Datasets}

\noindent \textbf{nuScenes dataset}~\cite{caesar2020nuscenes} contains 1000 scenes in total, split into 700 in the training set, 150 in the validation set, and 150 in the test set. Each sequence is captured at 20Hz frequency with 20 seconds duration.
Each sample contains RGB images from 6 cameras with 360$^{\circ}$ horizontal FOV and point cloud data from 32 beam LiDAR sensor.
For the task of object detection, the key samples are annotated at 2Hz with ground truth labels for 10 foreground object classes (\emph{thing}). For the task of semantic segmentation and panoptic segmentation, every point in the key samples is annotated using 6 more background classes (\emph{stuff}) in addition to the 10 foreground classes (\emph{thing}).

\definecolor{nbarrier}{RGB}{255, 120, 50}
\definecolor{nbicycle}{RGB}{255, 192, 203}
\definecolor{nbus}{RGB}{255, 255, 0}
\definecolor{ncar}{RGB}{0, 150, 245}
\definecolor{nconstruct}{RGB}{0, 255, 255}
\definecolor{nmotor}{RGB}{200, 180, 0}
\definecolor{npedestrian}{RGB}{255, 0, 255}
\definecolor{ntraffic}{RGB}{255, 240, 150}
\definecolor{ntrailer}{RGB}{135, 60, 0}
\definecolor{ntruck}{RGB}{255, 0, 0}
\definecolor{ndriveable}{RGB}{213, 213, 213}
\definecolor{nother}{RGB}{139, 137, 137}
\definecolor{nsidewalk}{RGB}{75, 0, 75}
\definecolor{nterrain}{RGB}{150, 240, 80}
\definecolor{nmanmade}{RGB}{160, 32, 240}
\definecolor{nvegetation}{RGB}{0, 175, 0}

\begin{table*}[ht]
	\footnotesize
 	\setlength{\tabcolsep}{0.0045\linewidth}
	
	\newcommand{\classfreq}[1]{{~\tiny(\nuscenesfreq{#1}\%)}}  %
    \begin{center}

	\begin{tabular}{l|c|c|c | c c c c c c c c c c c c c c c c}
		\toprule
		Method
		& \makecell{Input \\ Modality}& \makecell{Image \\ Backbone} & mIoU
		& \rotatebox{90}{\textcolor{nbarrier}{$\blacksquare$} barrier}
		
		& \rotatebox{90}{\textcolor{nbicycle}{$\blacksquare$} bicycle}
		
		& \rotatebox{90}{\textcolor{nbus}{$\blacksquare$} bus}

		& \rotatebox{90}{\textcolor{ncar}{$\blacksquare$} car}

		& \rotatebox{90}{\textcolor{nconstruct}{$\blacksquare$} const. veh.}

		& \rotatebox{90}{\textcolor{nmotor}{$\blacksquare$} motorcycle}

		& \rotatebox{90}{\textcolor{npedestrian}{$\blacksquare$} pedestrian}

		& \rotatebox{90}{\textcolor{ntraffic}{$\blacksquare$} traffic cone}

		& \rotatebox{90}{\textcolor{ntrailer}{$\blacksquare$} trailer}

		& \rotatebox{90}{\textcolor{ntruck}{$\blacksquare$} truck}

		& \rotatebox{90}{\textcolor{ndriveable}{$\blacksquare$} drive. suf.}

		& \rotatebox{90}{\textcolor{nother}{$\blacksquare$} other flat}

		& \rotatebox{90}{\textcolor{nsidewalk}{$\blacksquare$} sidewalk}

		& \rotatebox{90}{\textcolor{nterrain}{$\blacksquare$} terrain}

		& \rotatebox{90}{\textcolor{nmanmade}{$\blacksquare$} manmade}

		& \rotatebox{90}{\textcolor{nvegetation}{$\blacksquare$} vegetation}

		\\
		\midrule

        RangeNet++~\cite{milioto2019rangenet++} & LiDAR & - & 65.5 & 66.0 & 21.3 & 77.2 & 80.9 & 30.2 & 66.8 & 69.6 &  52.1 & 54.2 & {72.3} & {94.1} & 66.6 & 63.5 & 70.1 & 83.1 & 79.8 \\
		
		PolarNet~\cite{zhang2020polarnet} & LiDAR &- & 71.0 & 74.7 & 28.2 & 85.3 & 90.9 & 35.1 & 77.5 & 71.3 & 58.8 & 57.4 & 76.1 & 96.5 & 71.1 & 74.7 & 74.0 & 87.3 & 85.7  \\
		
		Salsanext~\cite{cortinhal2020salsanext} & LiDAR &- & 72.2 & 74.8 & 34.1 & 85.9 & 88.4 & 42.2 & 72.4 & 72.2 & 63.1 & 61.3 & 76.5 & 96.0 & 70.8 & 71.2 & 71.5 & 86.7 & 84.4 \\
		
		Cylinder3D++~\cite{zhu2021cylindrical} & LiDAR &- & 76.1 & 76.4 & 40.3 & 91.2 & 93.8 & 51.3 & 78.0 & 78.9 & 64.9 & 62.1 & 84.4 & 96.8 & 71.6 & 76.4 & 75.4 & 90.5 & 87.4 \\
        RPVNet~\cite{xu2021rpvnet} & LiDAR  &- & 77.6 & 78.2 & 43.4 & 92.7 & 93.2 & 49.0 & 85.7 & 80.5 & 66.0 & 66.9 & 84.0 & 96.9 & 73.5 & 75.9 & 76.0 & 90.6 & 88.9 \\
			\midrule	
		
        BEVFormer-Base~\cite{li2022bevformer} & Camera &R101-DCN & 56.2  & 54.0 & 22.8 & 76.7 & 74.0 & 45.8 & 53.1 & 44.5 & 24.7 & 54.7 & 65.5 & 88.5 & 58.1 & 50.5 & 52.8 & 71.0 & 63.0  \\
		
		TPVFormer-Base~\cite{huang2023tri}  & Camera & R101-DCN &68.9  & 70.0 & 40.9 & 93.7 & 85.6 & 49.8 & 68.4 & 59.7 & 38.2 & 65.3 & 83.0 & 93.3 & 64.4 & 64.3 & 64.5 & 81.6 & 79.3  \\ %
         \midrule
         \methodname-Base &  Camera & R101-DCN &70.7  & 73.7 & 42.6 & 94.1 & 87.1 & 56.4 & 62.4 & 64.7 & 36.7 & 69.3 & 86.4 & 94.9 & 69.8 & 67.1 & 67.9 & 80.3 & 77.0 \\ %

        \methodname-Small-T &  Camera & R50 &68.1 & 70.7 & 37.9 & 92.3 & 85.0 & 50.7 & 64.3 & 59.4 & 35.3 & 63.8 & 81.6 & 94.2 & 66.4 & 64.8 & 68.0 & 79.1 & 75.6 \\
        
         \methodname-Base-T &  Camera & R101-DCN& 71.6  & 74.3 &43.7  &95.4  &87.0  & 56.1 &64.6  &66.2  &41.4  &71.5 &85.9 &95.1 &70.1 &67.0  &68.1 &80.9 &77.4  \\ %

         \methodname-Large-T &  Camera& InternImage-XL &\textbf{74.5}  & \textbf{75.3 }& \textbf{51.1} &\textbf{ 96.9 }& \textbf{87.5} &\textbf{ 56.6} & \textbf{85.6} & \textbf{68.0} & \textbf{43.0} &\textbf{74.1} & \textbf{87.1}& \textbf{95.1}&\textbf{71.0} &\textbf{68.7}  &\textbf{70.3} & \textbf{82.3}&\textbf{79.3}  \\ %
  
		\bottomrule
	\end{tabular}
    \end{center}
    \caption{\textbf{LiDAR semantic segmentation results on nuScenes validation set.} Our \methodname{}-Large-T achieves comparable performance with state-of-the-art LiDAR-based methods. T denotes the usage of temporal information.
	}
 \vspace{-10pt}
	\label{tab:lidar_seg}
\end{table*} 

\noindent \textbf{Occ3D-nuScenes}~\cite{tian2023occ3d} contains 700 training scenes and 150 validation scenes. The occupancy scope is defined as $-40m$ to $40m$ for X and Y-axis, and $-1m$ to $5.4m$ for the Z-axis in the ego coordinate. The voxel size is $0.4m \times 0.4m \times 0.4m$ for the occupancy label. The semantic labels contain 17 categories (including `others'). Besides, it also provides visibility masks for LiDAR and camera modality.

\definecolor{nbarrier}{RGB}{255, 120, 50}
\definecolor{nbicycle}{RGB}{255, 192, 203}
\definecolor{nbus}{RGB}{255, 255, 0}
\definecolor{ncar}{RGB}{0, 150, 245}
\definecolor{nconstruct}{RGB}{0, 255, 255}
\definecolor{nmotor}{RGB}{200, 180, 0}
\definecolor{npedestrian}{RGB}{255, 0, 255}
\definecolor{ntraffic}{RGB}{255, 240, 150}
\definecolor{ntrailer}{RGB}{135, 60, 0}
\definecolor{ntruck}{RGB}{255, 0, 0}
\definecolor{ndriveable}{RGB}{213, 213, 213}
\definecolor{nother}{RGB}{139, 137, 137}
\definecolor{nsidewalk}{RGB}{75, 0, 75}
\definecolor{nterrain}{RGB}{150, 240, 80}
\definecolor{nmanmade}{RGB}{160, 32, 240}
\definecolor{nvegetation}{RGB}{0, 175, 0}
\definecolor{nothers}{RGB}{0, 0, 0}

\begin{table*}[ht]
	\footnotesize
 	\setlength{\tabcolsep}{0.0025\linewidth}
	
	\newcommand{\classfreq}[1]{{~\tiny(\nuscenesfreq{#1}\%)}}  %
    \begin{center}

	\begin{tabular}{l|c|c|c| c c c c c c c c c c c c c c c c c}
		\toprule
		Method
		& \makecell{Input \\ Modality}& \makecell{Image \\ Backbone} & mIoU
        & \rotatebox{90}{\textcolor{nothers}{$\blacksquare$} others}
        
		& \rotatebox{90}{\textcolor{nbarrier}{$\blacksquare$} barrier}
		
		& \rotatebox{90}{\textcolor{nbicycle}{$\blacksquare$} bicycle}
		
		& \rotatebox{90}{\textcolor{nbus}{$\blacksquare$} bus}

		& \rotatebox{90}{\textcolor{ncar}{$\blacksquare$} car}

		& \rotatebox{90}{\textcolor{nconstruct}{$\blacksquare$} const. veh.}

		& \rotatebox{90}{\textcolor{nmotor}{$\blacksquare$} motorcycle}

		& \rotatebox{90}{\textcolor{npedestrian}{$\blacksquare$} pedestrian}

		& \rotatebox{90}{\textcolor{ntraffic}{$\blacksquare$} traffic cone}

		& \rotatebox{90}{\textcolor{ntrailer}{$\blacksquare$} trailer}

		& \rotatebox{90}{\textcolor{ntruck}{$\blacksquare$} truck}

		& \rotatebox{90}{\textcolor{ndriveable}{$\blacksquare$} drive. suf.}

		& \rotatebox{90}{\textcolor{nother}{$\blacksquare$} other flat}

		& \rotatebox{90}{\textcolor{nsidewalk}{$\blacksquare$} sidewalk}

		& \rotatebox{90}{\textcolor{nterrain}{$\blacksquare$} terrain}

		& \rotatebox{90}{\textcolor{nmanmade}{$\blacksquare$} manmade}

		& \rotatebox{90}{\textcolor{nvegetation}{$\blacksquare$} vegetation}

		\\
		\midrule
        MonoScene~\cite{cao2022monoscene} & Camera & R101-DCN & 6.06 & 1.75 & 7.23 & 4.26 & 4.93 & 9.38 & 5.67 & 3.98 & 3.01 & 5.90 & 4.45 & 7.17 & 14.91 & 6.32 & 7.92 & 7.43 & 1.01 & 7.65\\
        BEVDet~\cite{huang2021bevdet} & Camera & R101-DCN & 11.73 & 2.09 & 15.29 & 0.0 & 4.18 & 12.97 & 1.35 & 0.0 & 0.43 & 0.13 & 6.59 & 6.66 & 52.72 & 19.04 & 26.45 & 21.78 & 14.51 & 15.26\\
		BEVFormer~\cite{li2022bevformer} & Camera & R101-DCN & 26.88 & 5.85 & 37.83 & 17.87 & 40.44 & 42.43 & 7.36 & 23.88 & 21.81 & 20.98 & 22.38 & 30.70 & 55.35 & 28.36 & 36.0 & 28.06 & 20.04 & 17.69 \\
        CTF-Occ~\cite{tian2023occ3d} & Camera & R101-DCN & 28.53 & 8.09 & 39.33 & 20.56 & 38.29 & 42.24 & 16.93 & 24.52 & 22.72 & 21.05 & 22.98 & 31.11 & 53.33 & 33.84 & 37.98 & 33.23 & 20.79 & 18.0\\
        BEVFormer*~\cite{li2022bevformer} & Camera & R101-DCN & 39.24 & 10.13 & 47.91 & 24.9 & 47.57 & 54.52 & 20.23 & 28.85 & 28.02 & 25.73 & 33.03 & 38.56 & 81.98 & 40.65 & 50.93 & 53.02 & 43.86& 37.15  \\
        BEVDet$\dagger$~\cite{huang2021bevdet} & Camera & Swin-B & 42.02 & 12.15 & 49.63 & 25.1 & 52.02 & 54.46 & 27.87 & 27.99 & 28.94 & 27.23 & 36.43 & 42.22 & 82.31 & 43.29 & 54.62 & 57.9 & 48.61 & 43.55  \\ 
        \midrule
        \methodname &  Camera & R101-DCN & \textbf{42.13} & 11.67 & 50.48 & 29.64 & 49.44 & 55.52 & 23.29 & 33.26 & 30.55 & 30.99 & 34.43 & 42.57 & 83.31 & 44.23 & 54.4 & 56.04 & 45.94 & 40.4  \\

		\bottomrule
	\end{tabular}
    \end{center}
    \caption{\textbf{3D Occupancy prediction performance on the Occ3D-nuScenes dataset.} $\dagger$ denotes the performance is reported by its official code implementation. * means the performance is achieved by our implementation using the camera mask during training.}
 \vspace{-10pt}
	\label{tab:camera_occ}
\end{table*}

\noindent \textbf{Evaluation metrics.} 
nuScenes dataset uses mean Average Precision (mAP) and nuScenes Detection Score (NDS) metrics for the detection task, mean Intersection over Union (mIoU), and Panoptic Quality (PQ) metrics~\cite{kirillov2019panoptic} for the semantic and panoptic segmentation.
PQ$^{\dagger}$ is a modified panoptic quality~\cite{porzi2019seamless}, which maintains the PQ metric for \emph{thing} classes, but modifies the metric for \emph{stuff} classes.
The Occ3D-nuScenes benchmark calculates the mean Intersection over Union (mIoU) for 17 semantic categories within the camera's visible region.

\subsection{Experimental Settings}
\noindent{\textbf{Implementation Details.}}
On the nuScenes dataset~\cite{caesar2020nuscenes}, we set the point cloud range for the $x$ and $y$ axes to $[-51.2m, 51.2m]$, and $[-5m, 3m]$ for the $z$ axis. The voxel grid size used for loss supervision is $(0.256m, 0.256m, 0.125m)$.
The initial resolution of the voxel queries is 50x50x16 for $H, W, Z$. We use an embedding dimension $D$ of 256, and learnable 3D position encoding is added to the voxel queries.
The upsampled voxel features have dimensions of 200x200x32 for $H', W', Z'$, and a feature dimension $D'$ of 64.
The backbone used in our approach includes ResNet50~\cite{he2016deep}, ResNet101-DCN~\cite{dai2017deformable}, and InternImage~\cite{wang2022internimage}, with output multi-scale features from FPN~\cite{lin2017feature} at sizes of 1/8,1/16,1/32 and 1/64.
The camera view encoder includes 3 layers, with each layer consisting of voxel self-attention, voxel cross-attention, norm layer, and feed-forward layer, with both $M_1$ and $M_2$ set to 4.
The temporal encoder fuses 4 frames (including the current frame) with a time interval of 0.5s. The voxel upsample module employs 3 layers of 3D deconvolutions to upscale 4x for $H$ and $W$, and 2x for $Z$, with detailed parameters in the supplementary materials.
The segmentation head has two MLP layers with a hidden dimension of 128 and uses \emph{softplus}~\cite{zheng2015improving} as the activation function. The number of object queries for the detection head is set to 900, and the loss weights used in our approach are $\lambda_1$=10.0, $\lambda_2$=10.0, $\lambda_3$=5.0, $\lambda_4$=2.0,and $\lambda_5$=0.25.

\noindent{\textbf{Training.}}
For the \methodname-Base setting, we adopted ResNet101-DCN~\cite{dai2017deformable} as the image backbone and trained the model on 8 NVIDIA A100 GPUs with a batch size of 1 per GPU. During training, we utilized the AdamW~\cite{loshchilov2017decoupled} optimizer for 24 epochs, with an initial learning rate of $2\times 10^{-4}$ and the cosine annealing schedule. Additionally, we employed several data augmentation techniques, including image scaling, color distortion, and Gridmask~\cite{chen2020gridmask}. The input image size is cropped to $640 \times 1600$.
For the \methodname-Large setting, we utilized InternImage-XL~\cite{wang2022internimage} as the image backbone, while the remaining training settings were the same as \methodname-Base.
For the \methodname-Small setting, we chose to use ResNet50~\cite{he2016deep} as the image backbone, and the input image size was resized to $320 \times 800$. During training, we did not utilize image scaling augmentation. 

\noindent{\textbf{Supervision.}}
For the detection head, we use object-level annotations as the supervision. 
We employ sparse LiDAR point-level semantic labels for the segmentation head to supervise voxel prediction.
When multiple semantic labels are present within a voxel grid, we prioritize the category label with the highest count of LiDAR points.

\noindent{\textbf{Evaluation.}}
Our approach can evaluate LiDAR semantic segmentation by assigning voxel semantic predictions to LiDAR points. We further extend it with object detection results, enabling panoptic voxel prediction and evaluation on the LiDAR panoptic segmentation benchmark~\cite{fong2022panoptic}.
As PQ is only computed on sparse points and cannot comprehensively reflect the understanding of foreground objects, we still choose to use mAP, NDS, and mIoU to measure the effectiveness of our approach in the experiments.

\subsection{Main Results}

\noindent{\textbf{3D Semantic Segmentation.}}
We assign the voxel predictions on sparse LiDAR points for the semantic segmentation evaluation. 
As shown in Table~\ref{tab:lidar_seg}, we evaluate the semantic segmentation performance on the nuScenes validation set.
\methodname-Base uses the ResNet101-DCN~\cite{dai2017deformable} initialized from FCOS3D~\cite{wang2021fcos3d} checkpoint.
\methodname-small adopts the ResNet-50~\cite{he2016deep} pretrained on ImageNet~\cite{deng2009imagenet}.
For a fair comparison, the setting of \methodname-Base is the same as TPVFormer-Base~\cite{huang2023tri}. 
Without bells and whistles, our \methodname-Base achieves an outstanding 70.7 mIoU, surpassing the previous state-of-the-art TPVFormer-Base~\cite{huang2023tri} by 1.8 points. 
Furthermore, by incorporating temporal information, \methodname-Base-T achieves an even higher mIoU score of 71.6.
To further validate our approach, we experimented with a larger image backbone~\cite{wang2022internimage}. As a result, \methodname-Large-T achieved an impressive 74.5 mIoU, comparable to the performance of current state-of-the-art LiDAR-based methods.

\noindent{\textbf{3D Occupancy Prediction.}} 
In Table~\ref{tab:camera_occ}, we present the evaluation results for 3D occupancy prediction on the Occ3D-nuScenes validation set. All methods utilize camera input and are trained for 24 epochs.
The performance of MonoScene~\cite{cao2022monoscene}, BEVDet~\cite{huang2021bevdet}, BEVFormer~\cite{li2022bevformer}, and CTF-Occ~\cite{tian2023occ3d} is reported in the work of Tian et al~\cite{tian2023occ3d}. 
The use of a camera visible mask during training has proven to be an effective technique. We re-implemented BEVFormer~\cite{li2022bevformer} with the inclusion of the camera mask during training. Similarly, BEVDet~\cite{huang2021bevdet} also adopts this trick and reports improved performance on its official code repository.
Our \methodname also use camera visibile mask during training and achieves a new state-of-art performance.
We adopt the R101-DCN backbone and use 4 frames for temporal fusion.

\noindent{\textbf{3D Panoptic Segmentation.}}
As \methodname{} is the first work to introduce camera-based panoptic segmentation, we can only compare it with previous LiDAR-based panoptic segmentation methods. The results in Table~\ref{tab:lidar_panoptic} show that our \methodname{} achieves 62.1 PQ, demonstrating comparable performance to some LiDAR-based methods such as EfficientLPS~\cite{sirohi2021efficientlps} and PolarNet~\cite{zhang2020polarnet}.
However, our approach still has a performance gap compared to state-of-the-art LIDAR-based methods, which can be attributed to our inferior detection performance (48.4 mAP v.s. 63.8 mAP).
\begin{table}[t]
    \small
    \begin{center}
     \setlength{\tabcolsep}{0.01\linewidth}
    \begin{tabular}{c|c|cccc|c}
		\toprule
		Method &  \makecell{Input \\ Modality} & PQ & PQ$^{\dagger}$ & RQ & SQ & mAP \\
		\midrule
        EfficientLPS~\cite{sirohi2021efficientlps} & LiDAR &62.0 & 65.6 & 73.9 & 83.4 & /\\
        Panoptic-PolarNet~\cite{zhou2021panoptic} & LiDAR & 63.4 & 67.2 & 75.3 &83.9 & /\\
        Panoptic-PHNet~\cite{li2022panoptic} & LiDAR & 74.7 & 77.7 & 84.2 & 88.2 & /\\
        LidarMulitiNet~\cite{ye2022lidarmultinet} & LiDAR & 81.8 & / & 90.8 & 89.7 & 63.8\\
        \midrule
        \methodname{}-Large-T & Camera& 62.1  &  66.2 & 75.1& 82.1 & 48.4 \\
		\bottomrule
	\end{tabular}
    \end{center}
    \caption{\textbf{LiDAR panoptic segmentation results on nuScenes validation set.} Our \methodname{} based on the camera input has approached LiDAR-based methods' performance.}
    \vspace{-10pt}
    \label{tab:lidar_panoptic}
    \end{table}

\subsection{Ablation}
We conduct ablation experiments on the design choices of \methodname{} on the nuScenes validation set. By default, we use the setting of \methodname{}-small.

\noindent{\textbf{Initial Voxel Resolution.}}
Table~\ref{tab:abl_resolution} compares the results of different initial resolutions used for voxel queries in our experiments. In experiments (b), (c), and (d), we maintained fixed dimensions of $H$ and $W$ while varying the resolution of $Z$. Our findings clearly demonstrate that encoding height information is a crucial factor in achieving superior performance in both segmentation(+5.3 mIoU) and detection tasks(+1.2 mAP and +1.6 NDS), with a more significant impact observed in segmentation tasks.
Furthermore, we observed that (a) and (b) have the same number of query parameters and perform similarly in detection tasks. However, there is a significant gap in the segmentation tasks between these two. Specifically, the mIoU gain from (d) to (a) is much less compared to that from (d) to (b). The experiment (e) results suggest that when the dimensions of $H$ and $W$ are too small, there will be a significant reduction in the performance of both detection and segmentation tasks.
Overall, our findings emphasize the importance of encoding height information to achieve fine-grained scene understanding.

\begin{table}[t]
    \small
    \begin{center}
     \setlength{\tabcolsep}{0.01\linewidth}
    \begin{tabular}{c|c|ccc}
		\toprule
		 & Query Resolution & mIoU & mAP & NDS  \\
		\midrule
        (a) & 100x100x4 & 0.617 & \textbf{0.276} & \textbf{0.327}\\
        (b) &50x50x16 & \textbf{0.661} & 0.271 & 0.324 \\
        (c) &50x50x8 & 0.631& 0.267 & 0.316 \\
         (d)&50x50x4 & 0.608 & 0.259 & 0.308 \\
        (e)&25x25x16 & 0.591 & 0.244 & 0.294\\
		\bottomrule
	\end{tabular}
 \end{center}
\caption{\textbf{Ablation study for different initial query resolutions.} Height information is important to achieve fine-grained 3D scene understanding.}
    \label{tab:abl_resolution}
\end{table}

\noindent{\textbf{Effectiveness of 3D Voxel Queries.}}
Table~\ref{tab:abl_vq} presents an ablation study where we investigate different query forms for queries. One common concern regarding 3D voxel queries is their computational complexity, which limits their usage at high resolutions. However, our results demonstrate that even at relatively small resolutions, voxel queries can still learn powerful representations, surpassing the performance of both 2D BEV queries and 2D Tri-plane queries.
It is worth noting that our comparison of query forms was conducted using the same parameter capacity, with each form consisting of approximately 40,000 queries.

\begin{table}[t]
    \small
    \begin{center}
     \setlength{\tabcolsep}{0.01\linewidth}
    \begin{tabular}{c|c|c|c}
		\toprule
		Method & Query form & Resolution & mIoU  \\
		\midrule
        BEVFormer-Base$^{*}$~\cite{li2022bevformer} & 2D BEV  & 200x200 & 56.2 \\
        TPVFormer-Base~\cite{huang2023tri} & 2D Tri-plane & 200x(200+16+16) & 68.8 \\
        \midrule
        \methodname{}-Base & 3D Voxel & 50x50x16 & \textbf{70.7 } \\
		\bottomrule
	\end{tabular}
    \end{center}
    \caption{\textbf{Ablation study for the query form design.}$^{*}$ represents the performance is implemented by~\cite{huang2023tri}. Base denotes the image backbone is ResNet101-DCN~\cite{dai2017deformable}.}
    \label{tab:abl_vq}
    \vspace{-10pt}
\end{table}

\noindent{\textbf{Efficiency of Coarse-to-Fine Design.}}
Table~\ref{tbl:efficiency} illustrates the advantages of our coarse-to-fine scheme, which utilizes a low-resolution 3D voxel grid. This approach not only helps in increasing performance and inference speed but also effectively reduces memory consumption. By comparing it with the direct use of high-resolution voxel queries (200x200x8), we observe that our coarse-to-fine design achieves comparable or even superior performance while consuming nearly half the memory. This demonstrates the efficiency and effectiveness of our approach.

\begin{table}[ht]
    \footnotesize
    \begin{center}
    \setlength{\tabcolsep}{0.01\linewidth}
      \begin{tabular}{c|c|cccc|c}
        \toprule
        \makecell{Voxel \\ Resolution} & \makecell{Voxel \\ Upsampling} & Memory & Latency & Param & FPS & mIoU  \\
        \midrule
        200x200x8& &37G / 9.5G & 255 ms & 117.7 M & 4.1  & 67.9\\
        50x50x16 &\checkmark&\textbf{18G} /\textbf{ 5.7G }& \textbf{149 ms} &\textbf{ 48.7 M}&\textbf{9.2}  &\textbf{68.3}\\
        \bottomrule
      \end{tabular}
    \end{center}
    \caption{\textbf{Ablation study for the coarse-to-fine design.} We show the train/inference memory consumption, respectively. The experiments were conducted on the A100 GPU.}
  \label{tbl:efficiency}
\end{table}

\noindent{\textbf{Design of Camera View Encoder.}}
Table~\ref{tab:abl_cam} presents the ablation study conducted on the design choices in the camera view encoder. Specifically, we experimented with different combinations of attention modules in (b), (c), and (d). The results demonstrated that incorporating voxel self-attention (VSA) enhanced the interaction between queries, leading to improved performance.
Considering both performance and parameters, we choose 3 layers as default.
\begin{table}[t]
    \small
    \begin{center}
        \setlength{\tabcolsep}{0.01\linewidth}
    \begin{tabular}{c|c|c|ccc}
		\toprule
		&Layers & Attention module & mIoU & mAP & NDS  \\
  		\midrule
        (a)&1 & VSA + VCA& 0.648 & 0.251 & 0.294\\
        (b)&3  & VCA  & 0.644 & 0.264 & 0.312\\   
        (c)&3  & VSA + VCA & 0.653 & 0.267 & 0.314 \\
        (d)&3  & VSA$\times$2 + VCA  & 0.661 & \textbf{0.271} & \textbf{0.324}  \\
        (e)&6  & VSA$\times$2 + VCA   & \textbf{0.662} & 0.267 & 0.319 \\
		\bottomrule
	\end{tabular}
    \end{center}
     \caption{\textbf{Ablation study for camera view encoder.} VSA denotes voxel self-attention, while VCA means voxel cross-attention.}
    \label{tab:abl_cam}
\end{table}

\noindent{\textbf{Design of Temporal Encoder.}}
Table~\ref{tab:abl_temporal} presents extensive ablation studies on the design of the temporal encoder, including different time intervals, number of frames, fusion methods, and encoder network architectures.
Compared to (a) and (b) designs, both detection and segmentation tasks show a significant improvement (+2.5 mIoU, +2.4 mAP, and +7.1 NDS), which suggests the importance of temporal information. 
In (b)(c)(d), we compared the influence of different time intervals and found that longer intervals do not improve the fine-grained segmentation performance. In (e) and (f), we also compared different ways to fuse the historical features and found that directly concatenating the features performs better than using temporal self-attention~\cite{li2022bevformer}.
\begin{table}[t]
    \small
    \begin{center}
        \setlength{\tabcolsep}{0.01\linewidth}  
    \begin{tabular}{c|c|c|c|c|c|ccc}
		\toprule
		&Temp. & Intv. & Frames & Fuse &Arch. & mIoU & mAP & NDS  \\
		\midrule
        (a)         & & / & 1 &/ & C3D$\times$1 & 0.656 & 0.269 & 0.319\\
      (b)&\checkmark & 0.5s & 4 & Cat. & C3D$\times$1 & \textbf{0.681} & 0.293 & \textbf{0.390} \\
      (c)&\checkmark & 1s & 4 & Cat. & C3D$\times$1 &  0.657 & \textbf{0.294} & 0.385\\
      (d)&\checkmark & 2s & 4 & Cat. & C3D$\times$1 &  0.660 &  0.294 & 0.375 \\
     (e)&\checkmark & 1s & 4 & Cat. & C3D$\times$3 &  0.658 & 0.290 & 0.379\\
      (f)&\checkmark & 0.5 & 4 & TSA & DA & 0.648 & 0.271 & 0.323 \\
		\bottomrule
	\end{tabular}
    \end{center}
     \caption{\textbf{Ablation study for temporal encoder.}
     Temp. stands for temporal fusion, while \checkmark denotes using temporal fusion. Intv. denotes time interval. Arch. refers to the architecture used in temporal encoder. C3D represents 3D convolution. $\times3$ means using 3 blocks of the architecture. Cat. means concatenating features from different frames, and TSA represents the temporal self-attention structure in ~\cite{li2022bevformer}. DA means deformable attention~\cite{zhu2020deformable}.}
    \label{tab:abl_temporal}
\end{table}

\begin{figure*}[ht]
          \begin{center}
              \includegraphics[width=\linewidth]{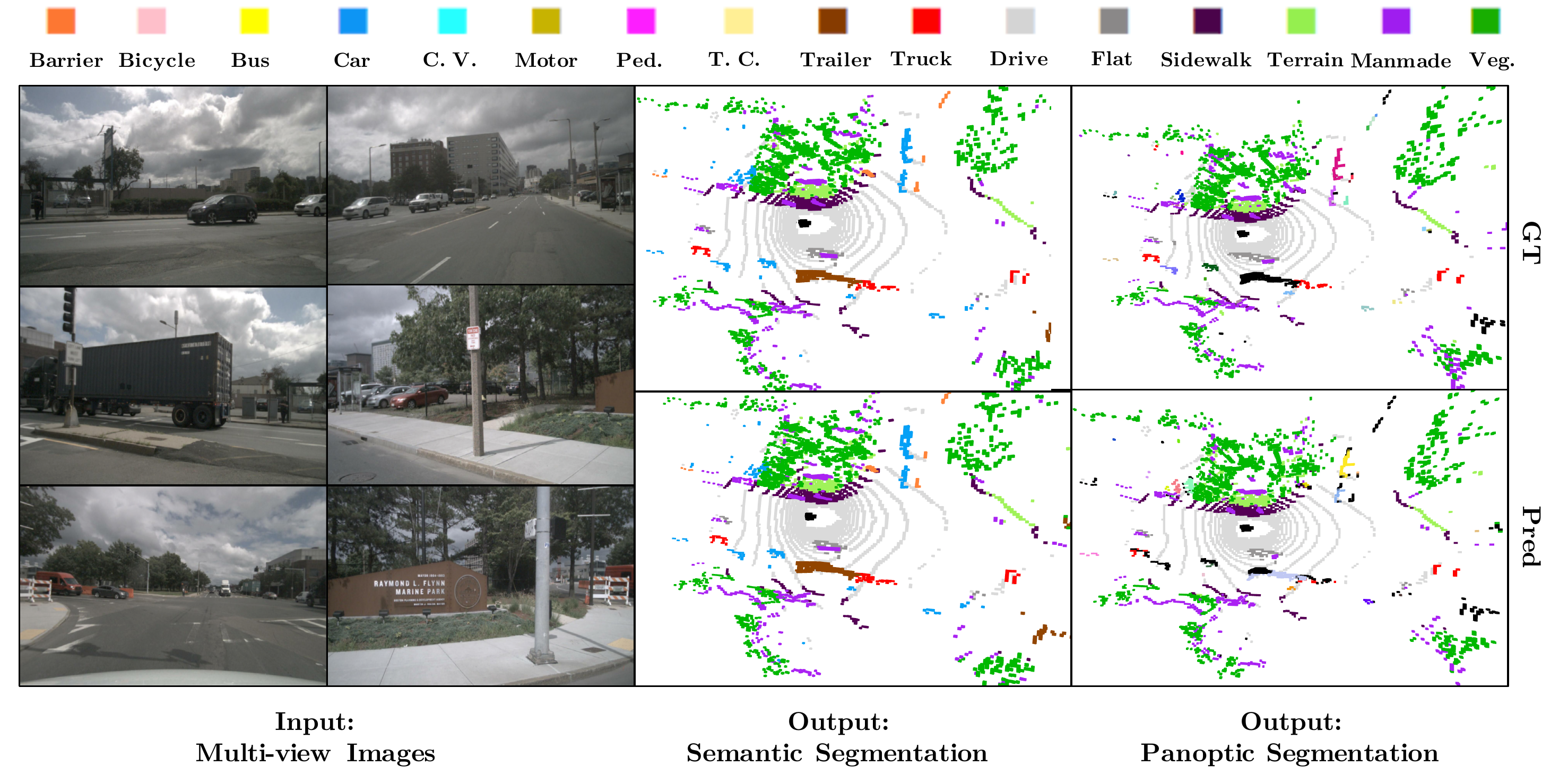}
          \end{center}
          \vspace{-12pt}
             \caption{\textbf{Qualitative results on nuScenes validation set.} Our \methodname{} takes multi-view images as input and produces voxel predictions, which are visualized at a resolution of 200x200x32. We evaluate 3D segmentic segmentation and panoptic segmentation on LiDAR points.}
             \label{fig:vis}
            \vspace{-10pt}
\end{figure*} 

\begin{figure*}[ht]
          \begin{center}
              \includegraphics[width=\linewidth]{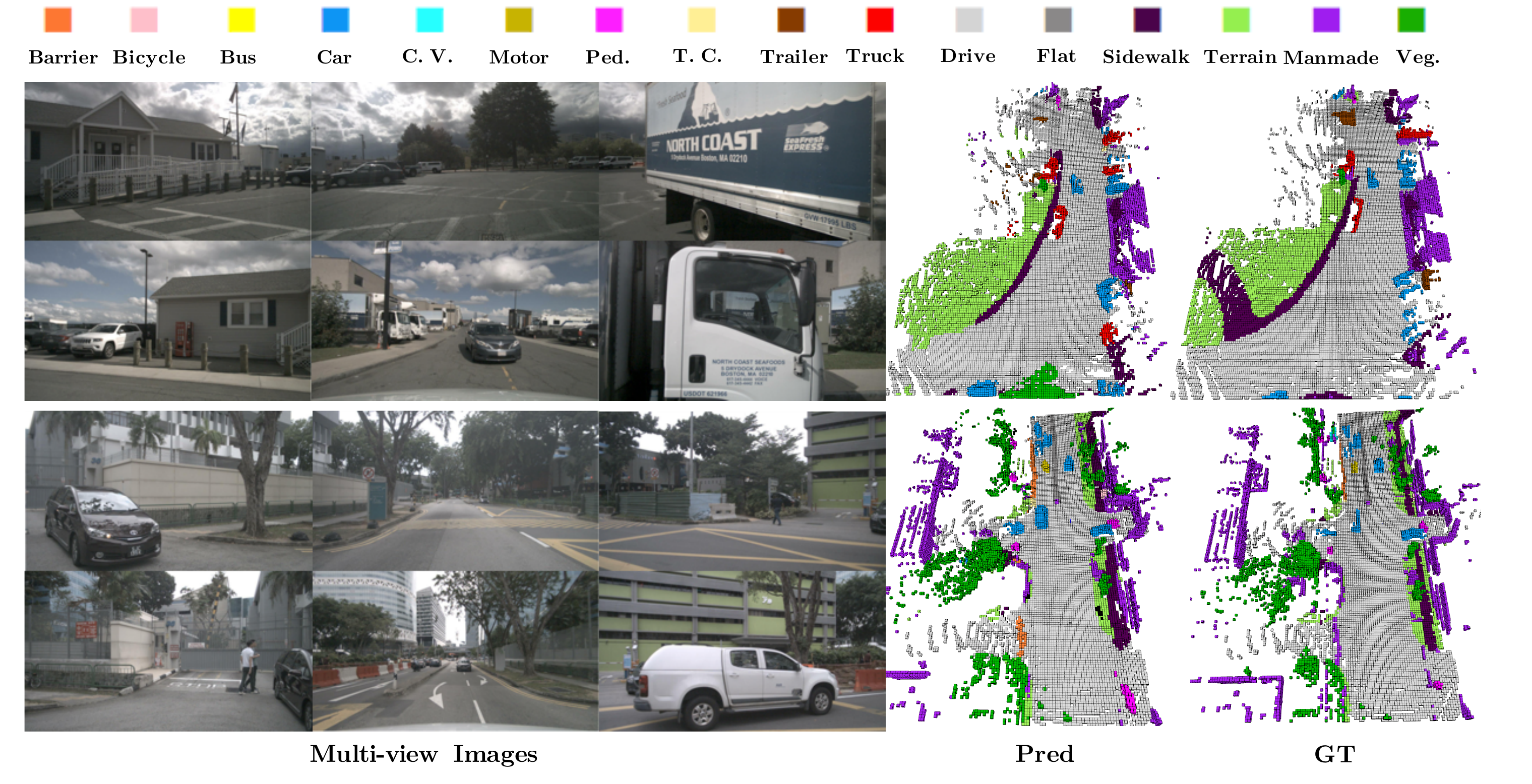}
          \end{center}
          \vspace{-12pt}
             \caption{\textbf{Qualitative results on Occ3D-nuScenes validation set.} Our \methodname{} takes multi-view images as input and produces dense occupancy predictions, which are visualized at the resolution of 200x200x16.}
             \label{fig:vis_dense}
            \vspace{-10pt}
\end{figure*} 

\noindent{\textbf{Effect of Joint Detection and Segmentation.}}
Table~\ref{tab:abl_joint} verifies the positive effect of joint detection and segmentation. In comparison to single-task models, the jointly-trained model performs better in both segmentation and detection tasks. 
Voxel selection further enhances the interaction between detection and segmentation learning, improving performance in both tasks. The unified voxel representation also enables efficient training by sharing the learning process of voxel features.

\begin{table}[t]
    \small
    \begin{center}

    \begin{tabular}{c|ccc|ccc}
		\toprule
		&Det. & Seg. & Vox. Sel. & mIoU & mAP & NDS\\
		\midrule
        (a)&\checkmark  &  & &  / & 0.252 & 0.310 \\
        (b)& &\checkmark & & 0.652&  /& / \\
       (c)& \checkmark &\checkmark &  & 0.656& 0.266 &0.319  \\
        (d)&\checkmark &\checkmark & \checkmark & \textbf{0.661} & \textbf{0.271} & \textbf{0.324} \\
		\bottomrule
	\end{tabular}
    \end{center}
     \caption{\textbf{Effect of joint detection and segmentation.} Det. means detection head. Seg. denotes segmentation head. Vox. Sel. represents voxel selection.}
    \label{tab:abl_joint}
    \vspace{-10pt}
\end{table}

\noindent{\textbf{The Supervision for Voxel Representation.}}
Table~\ref{tab:abl_sup} ablates the effects of different resolutions for segmentation loss supervision.
The experiment results indicate that resolution at 400x400x64 has the best performance.
\begin{table*}[ht]
	\footnotesize
 	\setlength{\tabcolsep}{0.01\linewidth}
  \begin{center}

  \begin{tabular}{c|cccccccccccccccc}
		\toprule
        mIoU
		& \rotatebox{90}{barrier}
		
		& \rotatebox{90}{bicycle}
		
		& \rotatebox{90}{bus}

		& \rotatebox{90}{car}

		& \rotatebox{90}{const. veh.}

		& \rotatebox{90}{motorcycle}

		& \rotatebox{90}{pedestrian}

		& \rotatebox{90}{traffic cone}

		& \rotatebox{90}{trailer}

		& \rotatebox{90}{truck}

		& \rotatebox{90}{drive. suf.}

		& \rotatebox{90}{other flat}

		& \rotatebox{90}{sidewalk}

		& \rotatebox{90}{terrain}

		& \rotatebox{90}{manmade}

		& \rotatebox{90}{vegetation}  \\
		\midrule
         65.6 & 72.3 & 35.8 & 91.4 &84.4  &47.2  &52.6  &57.7 &31.5  &55.6  &80.6  &94.0 &64.3  &63.2 &66.5 &77.7  &73.9  \\
        \makecell{68.1 \\ (\textcolor{green}{2.5$\uparrow$})} & \makecell{70.7 \\(\textcolor{red}{1.6$\downarrow$})} &  \makecell{37.9 \\(\textcolor{green}{2.1$\uparrow$})} &  \makecell{92.3\\(\textcolor{green}{0.9$\uparrow$})} &  \makecell{85.0\\(\textcolor{green}{0.6$\uparrow$})} &  \makecell{50.7\\(\textcolor{green}{3.5$\uparrow$})} & \makecell{64.3\\ (\textcolor{green}{11.7$\uparrow$})}& \makecell{59.4\\(\textcolor{green}{1.7$\uparrow$})} & \makecell{35.3\\(\textcolor{green}{3.8$\uparrow$})} & \makecell{63.8\\(\textcolor{green}{8.2$\uparrow$})} & \makecell{81.6\\(\textcolor{green}{1.0$\uparrow$})} & \makecell{94.2\\(\textcolor{green}{0.2$\uparrow$})} & \makecell{66.4\\(\textcolor{green}{2.1$\uparrow$})} & \makecell{64.8\\(\textcolor{green}{1.6$\uparrow$})} & \makecell{68.0\\(\textcolor{green}{1.5$\uparrow$})} & \makecell{79.1\\(\textcolor{green}{1.4$\uparrow$})} & \makecell{75.6\\(\textcolor{green}{1.7$\uparrow$})} \\
		\bottomrule
	\end{tabular}
   \end{center}
     \caption{\textbf{Effect of temporal enhancement on different categories.} The findings indicated that incorporating temporal information improved segmentation performance for most categories. We use the setting of \methodname-Small in the ablation.}
    \label{tab:abl_category}    
    \vspace{-10pt}
\end{table*}
\begin{table}[t]
    \small
    \begin{center}
        \setlength{\tabcolsep}{0.01\linewidth}
    
    \begin{tabular}{c|c|c|ccc}
		\toprule
		Supervision & Voxel feats& Loss Resolution &  mIoU & mAP & NDS  \\
		\midrule
        LiDAR & 200x200x32 & 400x400x64 &  \textbf{0.661} & \textbf{0.271} & \textbf{0.324} \\
        LiDAR & 200x200x32 & 200x200x32  &0.644 & 0.267 & 0.316\\
        LiDAR &100x100x16 & 100x100x16 & 0.609  & 0.264 & 0.317 \\
		\bottomrule
	\end{tabular}
    \end{center}
    \vspace{-5pt}
     \caption{\textbf{Supervision for voxel representation.} We utilize sparse LiDAR point labels as the supervision for voxel representation.}
    \label{tab:abl_sup}
\end{table}
\vspace{5pt}
\begin{table}[t]
    \small
    \setlength{\tabcolsep}{0.01\linewidth}
    \begin{center}
    \begin{tabular}{ccc|ccc|ccc}
		\toprule
		$\mathcal{L}_{focal}$&$\mathcal{L}_{lovasz}$& $\mathcal{L}_{thing}$
        &$\lambda_1$ & $\lambda_2$  &$\lambda_3$ &  mIoU & mAP & NDS  \\
		\midrule
        \checkmark & & &10.0 & / & / & 0.596 & 0.259 & 0.315\\
        \checkmark & \checkmark& &10.0 & 10.0 & / & 0.656 & 0.266 & 0.319 \\
         & \checkmark& \checkmark &/ & 10.0 & 5.0 & 0.643 & 0.260 & 0.311\\
        \midrule
        \checkmark &\checkmark &\checkmark & 10.0 & 10.0 & 5.0 &\textbf{ 0.661} & \textbf{0.271} & \textbf{0.324}  \\
        \checkmark &\checkmark &\checkmark & 10.0 & 10.0 & 10.0 & 0.652 & 0.265 & 0.317 \\
        \checkmark &\checkmark &\checkmark & 5.0 & 10.0 & 5.0 &0.656 & 0.266 & 0.315 \\
        \checkmark &\checkmark &\checkmark & 15.0 & 10.0 & 5.0&0.650  &0.265 & 0.314 \\
        \checkmark &\checkmark &\checkmark & 10.0 & 15.0 & 5.0 & 0.654 & 0.263 & 0.312 \\
		\bottomrule
	\end{tabular}
    \end{center}
        \vspace{-5pt}
     \caption{\textbf{Ablation for loss terms and weights.} We ablates different loss combinations and its weight.}
    \label{tab:abl_loss}
     % \vspace{-10pt}
\end{table}
\vspace{5pt}

\noindent{\textbf{Loss Terms and Weights.}}
Table~\ref{tab:abl_loss} presents the comparison of various combinations of loss terms and weights.
It indicates that the $\mathcal{L}_{lovasz}$ plays a crucial role in the segmentation learning process, as its removal led to a significant drop in performance (from 65.6 to 59.6 mIoU). We also experimented with various weight combinations and found that $\lambda_1=10, \lambda_2=10, \lambda_3=5$ performs best.

\begin{table}[ht]
    \small
    \begin{center}
      \begin{tabular}{c|ccc|c}
        \hline
        \multicolumn{1}{c|}{Method}  &Memory & Latency & FPS & mIoU  \\
        \hline
        TPVFormer-Base$^*$ & 33.5G / 7.1G & 268 ms & 3.7  & 68.9\\
        \methodname-Base &\textbf{24G} / \textbf{6.0G}& \textbf{203 ms} &  \textbf{4.8} & \textbf{71.7}\\
        \hline
      \end{tabular}
          \end{center}
          \vspace{-5pt}
    \caption{\textbf{Model efficiency comparison}. $^*$ denotes the performance using its official code and released checkpoints. We report the train/inference memory consumption in the experiment.}
  \label{tbl:compare}
\end{table}
\vspace{5pt}
\begin{table}[t]
    \small
    \begin{center}
     \setlength{\tabcolsep}{0.015\linewidth}
    \begin{tabular}{c|c|ccc|c}
		\toprule
		 & Convolution & Latency &Memory & FPS & mIoU   \\
		\midrule
        (a) & Dense & 126 ms & 15 G& 9.3 &\textbf{0.654} \\
        (b) & Sparse & \textbf{112 ms} & \textbf{9 G}&\textbf{ 9.7} & 0.639\\
		\bottomrule
	\end{tabular}
 \end{center}
\caption{\textbf{Exploration of sparse architecture design.} The experiment is conducted under the \methodname-small setting without temporal fusion.}
    \label{tab:sparse}
    \vspace{-10pt}
\end{table}

\subsection{Discussion}

\noindent{\textbf{Voxel vs. Tri-plane.}} 
Traditionally, it has been believed that using 3D voxel grids alone is an inefficient solution due to the memory cost. This has led methods like TPVFormer~\cite{huang2023tri} to split the 3D representation into three 2D planes. 
However, we have demonstrated for the first time that using the coarse-to-fine voxel representation can solve the memory increasing problem.
In Table~\ref{tbl:compare}, we compare the performance and efficiency of our method with the previous state-of-the-art approach, TPVFormer~\cite{huang2023tri}, under the same experimental setup. Despite having an additional detection branch and the capability to output detection results, our model still exhibits lower memory consumption and faster inference speed.

\noindent{\textbf{Occupancy Sparsify.}}
In contrast to 2D space, 3D space exhibits high sparsity, indicating that the majority of voxels are empty. In Table~\ref{tab:sparse}, we investigate the effectiveness of the occupancy sparsify strategy. Here we have 3 layers of sparse deconvolution for upsampling in total. In coarse-to-fine order, the keeping ratio after each upsampling is 0.2, 0.5, and 0.5, respectively.
It suggests that finally we only keep 5\% voxels.

\noindent{\textbf{Temporal Enhancement.}}
In Table~\ref{tab:abl_category}, we compared the impact of temporal information on different categories. The findings revealed that the semantic segmentation performance improved for almost all categories except for the barrier category. The motorcycle and trailer categories demonstrated a significant improvement, with a boost of 11.7 mIoU and 8.2 mIoU, respectively. These two categories are typically affected by occlusion, and thus, the utilization of temporal information can enhance the model's ability to accurately detect and segment occluded objects.

\subsection{Visualization}
Figure~\ref{fig:vis} showcases qualitative results achieved by \methodname{} on the nuScenes validation set. The voxel predictions are visualized at a resolution of 200x200x32 and assign to LiDAR points. These visualizations highlight the accuracy and reliability of our predictions for 3D semantic segmentation and panoptic segmentation.
Figure~\ref{fig:vis_dense} illustrates the dense occupancy prediction on the Occ3D-nuScenes validation set, where voxel predictions are visualized at the resolution of 200x200x16.

%-------------------------------------------------------------------------

\section{Conclusion}

In this paper, we propose \emph{camera-based 3D panoptic segmentation}, aiming for a comprehensive understanding of the scene by a unified occupancy representation.
To facilitate occupancy representation learning, we propose a novel framework called \methodname{} that utilizes voxel queries to incorporate information from multi-frame and multi-view images in a coarse-to-fine scheme.
Extensive experiments on the nuScenes dataset and Occ3D-nuScenes demonstrate the effectiveness of \methodname{} and its potential to advance holistic 3D scene understanding. We envision 3D occupancy representation as a promising new paradigm for future 3D scene perception.

\clearpage

{\small
\bibliographystyle{ieee_fullname}
\bibliography{egbib}
}

\end{document}